\documentclass[runningheads]{llncs}

\usepackage[T1]{fontenc}
\usepackage[utf8]{inputenc}
\usepackage{graphicx}
\usepackage{amsmath}
\usepackage{amssymb}
\usepackage{booktabs}
\usepackage{float}
\newcommand{\blfootnote}[1]{%
  \begingroup
  \renewcommand{\thefootnote}{}%
  \footnotetext{#1}%
  \addtocounter{footnote}{-1}%
  \endgroup
}

\begin{document}

\title{Hierarchical Specialised Ensembles for Classification of Zebrafish Phenotypes Using the Selected Image Recognition Methods}
\titlerunning{Hierarchical Specialised Ensembles for Zebrafish Phenotypes}

\author{Piotr S. Maci\k{a}g\inst{1}\thanks{Corresponding author.} \and
Monika Maci\k{a}g\inst{2,3} \and
Magdalena Majdan\inst{4}}
\authorrunning{P. S. Maci\k{a}g et al.}

\institute{Institute of Computer Science, Warsaw University of Technology,
Nowowiejska 15/19, 00-665 Warsaw, Poland\\
\email{piotr.maciag@pw.edu.pl}
\and
Laboratory of Neuronal Plasticity, Nencki Institute of Experimental Biology
of the Polish Academy of Sciences, Warsaw, Poland
\and
Laboratory of Behavioral Studies, Medical University of Lublin,
Lublin, Poland
\and
Department of Toxicology and Food Science, Medical University of Warsaw,
Banacha 1, 02-097 Warsaw, Poland}

\maketitle

\begin{abstract}
We propose and evaluate three hierarchical ensemble setups for zebrafish phenotype classification from embryo images. In all setups, stage 1 uses a single four-class classifier to assign images to one of the exclusive phenotypes: Normal, Chorion, Dead, or Other. Images classified as Other are then processed in stage 2, where the ensemble design differs across setups: a single multi-label classifier, two specialized multi-label classifiers, or an ensemble of binary classifiers. We compare these setups using three backbone architectures: ResNet18, ViT, and ConvNeXt. Overall, ConvNeXt achieves the best performance across setups, while the specialized hierarchical ensemble in setup 2 provides the best balance in terms of F1-score. The results show that the proposed specialised hierarchical ensembles are effective for zebrafish phenotype recognition, and suggest that ConvNeXt is particularly useful backbone model.

\keywords{Ensemble machine learning \and Zebrafish image \and Deep neural networks}
\end{abstract}

\section{Introduction}
\label{sec:Intro}
\blfootnote{Accepted manuscript for publication in Procedia Computer Science,
following presentation at KES 2026. This manuscript version is licensed under
CC BY-NC-ND 4.0.}
Testing and assessment of various toxicological substances play an important role in pharmacological research. Such assessments are commonly conducted using zebrafish (\textit{Danio rerio}) larvae due to their rapid development and high susceptibility to the effects of chemical exposure. Zebrafish share approximately 70\% of their genome with other mammalian species \cite{Tandon2025-AIToxicityZebrafish,Gunnarsson2008-ZebrafishMammals}, and possess organs and physiological functions analogous to those found in humans \cite{MacRae2015-Zebrafish}. Consequently, zebrafish larvae have become a widely used model organism for studying the effects of various pharmacological substances, e.g., quinoline yellow \cite{Majdan2025-ZebrafishQuinoline} or various types of metal particles \cite{Haque2018-ZebrafishSubstances,Maciag2022-ZebrafishCardiotoxicity,Maciag2022-BetaAdrenergicZebrafish}.

Despite these advantages, the manual assessment of morphological alterations in microscopy images of zebrafish larvae remains time-consuming, subjective, and prone to inter-observer variability, highlighting the need for more robust and automated analytical approaches.  Consequently, many researchers have attempted to develop machine learning pipelines capable of performing automated analysis of such images. These approaches aim, for example, to identify the developmental or physiological state of larvae and to perform accurate segmentation of microscopic images \cite{Tyagi2018-CNNsZebrafish,Wlodkowic2022-High-throughput}. Specific machine learning models involved in the past are, for example, extremely randomized trees \cite{Jeanray2015-phenotype} or various types of deep neural networks \cite{Shang2020-zebrafishcnns,Tandon2025-AIToxicityZebrafish}.

In this study, we experimentally compare effectiveness of image classification methods applied for recognition of zebrafish phenotypes. To this end, we propose three hierarchical ensemble setups, each consisting of two stages. 

The three proposed ensemble setups are tested by integrating three backbone methods\footnote{The implementations of all three backbone methods are taken from the \textit{timm} Python package.}:
\begin{itemize}
    \item A traditional Convolutional Neural Network (CNN) based on a pretrained ResNet18 model. CNNs were previously tested in classification of zebrafish phenotypes (see, for example, \cite{Jeanray2015-phenotype,Majdan2025-ZebrafishQuinoline}). We fine-tuned a pretrained ResNet18 with images of zebrafish by adding a single classification layer on top of the rest of ResNet18 layers.
    
    \item Vision Transformers (ViT), which adapt a Transformer architecture to visual tasks, such as image recognition, segmentation or generation. The ViT architecture is based on a mechanism called multi-head self-attention, which is taught using sequences of images' "patches" flattened to numerical vectors \cite{Khan2022-ViTs}. Generally, it has been shown that ViT can outperform CNNs (even those accompanied by a single-head attention mechanism) and can be successfully applied to zebrafish phenotypes classification \cite{Puchalla2025Zebrafish}.
    
    \item The ConvNeXt architecture, which combines advantages of CNNs and ViTs \cite{Liu2022-convnext}, but does not directly use any attention mechanisms (like CBAM \cite{Woo2018-cbam}). ConvNeXt was designed by iteratively incorporating to CNNs all design choices previously used in a ViT architecture, such as "patchifying" layer or larger kernel sizes. To the best of our knowledge, the ConvNeXt architecture was not previously used to classify zebrafish phenotypes. 

\end{itemize}

These three backbone modes are employed by us in three hierarchical ensemble setups explained in Section~\ref{sec:PM}.

\subsection{Contributions}

The two main contributions of our work are:
\begin{enumerate}
    \item Overall, the ConvNeXt architecture yields the best classification results, even for the most difficult phenotypes to classify, such as Downed Curved Tail. This is a notable finding, as the ConvNeXt architecture has not been previously explored in the context of zebrafish phenotype classification.
    
    \item The proposed hierarchical specialised ensemble (our proposed setup~2) tends to perform best among three tested setups, especially for ConvNeXt and ViT backbones. In contrast, prior studies (e.g., Jeanray et al.~\cite{Jeanray2015-phenotype} and Tandon et al.~\cite{Tandon2025-AIToxicityZebrafish}) mainly employ approaches using specialised binary classifiers. Our finding indicate that creating either too general or too specialized ensemble may worsen the achieved classification quality. 
\end{enumerate}

\subsection{Structure}

The structure of the article is as follows: in Section~\ref{sec:RW}, we review the related work. We specifically focus in advancements in development of machine learning approaches to classification of morphological changes in zebrafish embryos. Section~\ref{sec:Dataset} describes the features and preparing of the used dataset. In Section~\ref{sec:PM}, we offer the proposed methodology and explain the experimental setup: used tools, environment and models. In Section~\ref{sec:Results}, we discuss the results of experiments and in Section~\ref{sec:Conclusions} we conclude the work.

\section{Related Work}
\label{sec:RW}

Early, Vogt et al. \cite{Vogt2009-Automated} developed an automatic approach to segment images of zebrafish embryos using Cognition Network Technology (CNT). Specifically, they approach showed that such segmentation can be successfully applied to identify changing phenotypes in embryos (such as yolk detection). 

Jeanray et al.~\cite{Jeanray2015-phenotype} applied extremely randomized trees to classify several phenotypes of zebrafish larvae, including Dead, Hemostasis, Edema, and Short Tail. In their study, zebrafish embryos were exposed to various chemical compounds, such as valproic acid and heavy metals. The acquired microscopic images were first preprocessed and manually annotated into 11 phenotype categories. Subsequently, a supervised learning approach based on an extremely randomized trees classifier was employed to automate the classification process. The model was trained on a labeled training dataset and then evaluated on a separate test dataset.

The proposed model achieved high recall in identifying most phenotypes in the test set, particularly Dead, Necrosed Yolk Sac, and Chorion. In our study, we used the public dataset provided by Jeanray et al.~\cite{Jeanray2015-phenotype}. The dataset consists of high-quality images of zebrafish representing 11 phenotypes. Among them, three phenotypes: Dead, Chorion, and Normal are mutually exclusive, while the remaining phenotypes may co-occur within a single image. The dataset was originally split into training and testing subsets by the authors.

The phenotype classification approach proposed by Jeanray et al. follows a two-stage procedure. In the first stage, a classifier identifies one of three mutually exclusive states present in an embryo image: Dead, Chorion, or Other. In the second stage, a one-vs-all scheme is applied to detect co-occurring defects (such as Hemostasis, Edema, or Short Tail) in images that were classified as Other in the first stage. In this scheme, each defect is assigned a dedicated binary classifier that determines whether the defect is present in a given image. Consequently, the training dataset for each binary classifier consists of positive examples (images exhibiting the given defect) and negative examples (images belonging to all other classes). 

The approach proposed by Jeanray et al.~\cite{Jeanray2015-phenotype} appears to exhibit two methodological limitations. First, it emphasizes maximizing recall (i.e., detection rate) in image classification, while not reporting the corresponding precision, which limits the assessment of the model's overall performance. Second, in the second-stage binary classification, the construction of training datasets introduces potential ambiguity: the same image may be included in both positive and negative classes across different classifiers. This occurs because a single image can simultaneously exhibit multiple defects (e.g., Hemostasis and Edema), leading to overlapping class assignments within the one-vs- all scheme. While we use the dataset of Jeanray et al.~\cite{Jeanray2015-phenotype} in our study, in the methodology section we explained how we avoided these limitations. 

In their work, Taygi et al. \cite{Tyagi2018-CNNsZebrafish} applied CNNs to classify zebrafish images from the dataset used in Jeanray et al.~\cite{Jeanray2015-phenotype}. First, they employed a five-layer CNN trained from scratch (that means, the network was trained solely using training dataset of zebrafish images). Second, they applied a similar from-scratch training approach using the VGG16 architecture. Finally, they conducted an experiment in which they fine-tuned a pretrained VGG16 model. A limitation of this approach is that it uses single-label image classification, which contradicts the above-explained design of the dataset, in which a the same image is often assigned multiply co-occurring labels by the authors.

Other approaches explored in the literature include the use of a multi-view classification scheme, in which a zebrafish embryo is imaged from two perspectives: dorsal and lateral. A CNN is then trained simultaneously on images from both views. The results suggest that this approach yields better classification performance than a single-view approach \cite{Tandon2025-AIToxicityZebrafish}. Puchalla et al. \cite{Puchalla2025Zebrafish} investigated a rolling-window approach to the classification of zebrafish phenotypes, in which images of the same embryo collected at successive time points (days) were used to train both a CNN and a Vision Transformer (ViT) model. As noted by the authors, this approach reduces the number of embryos required by augmenting the training dataset with images capturing different developmental stages of the same individuals. Thus, to maintain a high identification recall of embryotypes, it is sufficient to train the model on images capturing the evolving states of the same individuals, rather than acquiring images from additional specimens.

\section{Used Dataset}
\label{sec:Dataset}

As noted above, our experiments were conducted using the dataset previously introduced by Jeanray et al.~\cite{Jeanray2015-phenotype}. The dataset comprises images representing 11 distinct phenotypes and is partitioned into training and testing subsets. From the training subset, approximately 20\% of the data was further separated in a stratified manner to form a validation set. These three splits were kept consistent across all evaluated model configurations. Among the 11 phenotypes, three are mutually exclusive: Chorion, Dead, and Normal\footnote{In the testing dataset, two images labeled as Normal are additionally annotated as either Edema or Hemostasis; however, we consider this inconsistency negligible.}. The remaining phenotypes frequently co-occur within the same images. These include Down Curved Tail, Edema, Hemostasis, Necrosed Yolk Sac, Short Tail, Up Curved Fish, Up Curved Tail, and Up Curved TailFish.

\begin{figure}
\centerline{\includegraphics[width=1\textwidth]{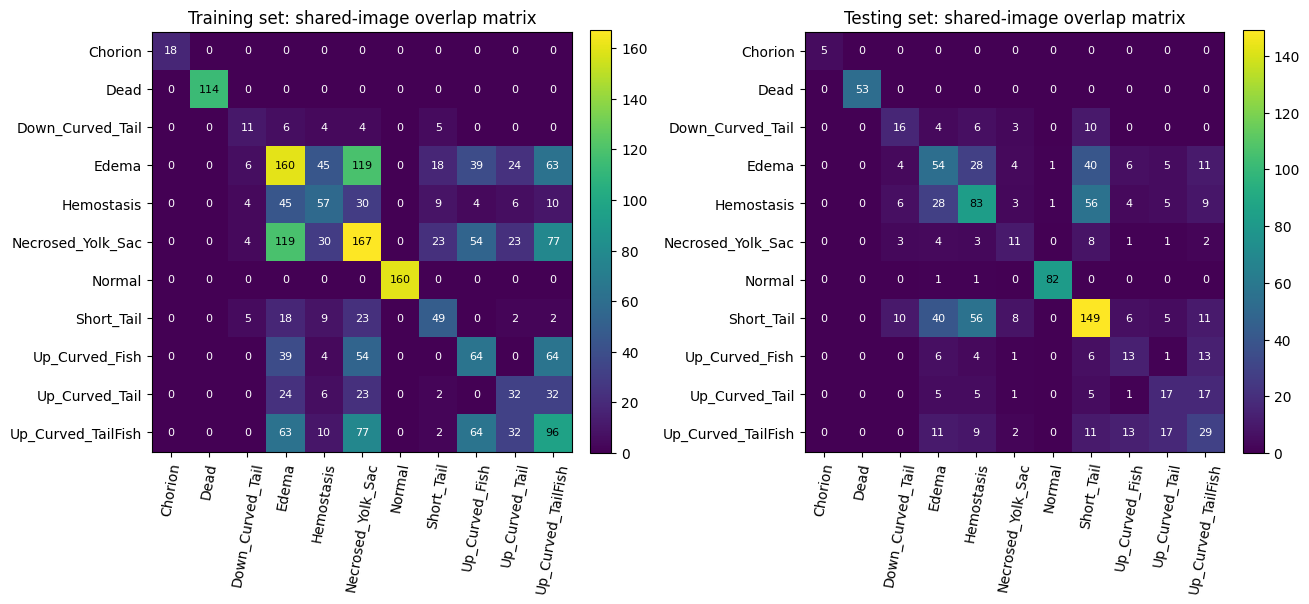}}
\caption{The heatmaps present the number of overlapping images for each class and dataset: training (left heatmap) and testing (right heatmap).}
\label{fig:heatmap}
\end{figure}

Fig.~\ref{fig:heatmap} presents heatmaps for both the training dataset (prior to validation split) and the testing dataset, illustrating the degree of overlap between phenotypes. These heatmaps were generated by computing and comparing SHA256 hashes of images across phenotype labels. The presence of multi-label annotations indicates the need to explicitly account for label co-occurrence when training image classification models. Overall, after extracting the validation subset, the training dataset consists of 742 images, the validation dataset contains 186 images, and the testing dataset comprises 512 images.

Additionally, we applied the following transformations to images:
\begin{itemize}
    \item For training/validation images: resizing to 224x224 pixels, random horizontal flip, random rotation by 10 degrees, scaling with \textit{ToTensor} function, normalization.
    \item For testing images: resizing to 224x224 pixels,  scaling with \textit{ToTensor} function, normalization. 
\end{itemize}

\section{Methodology}
\label{sec:PM}

\begin{figure}[h!t]
\centerline{\includegraphics[width=0.9\textwidth]{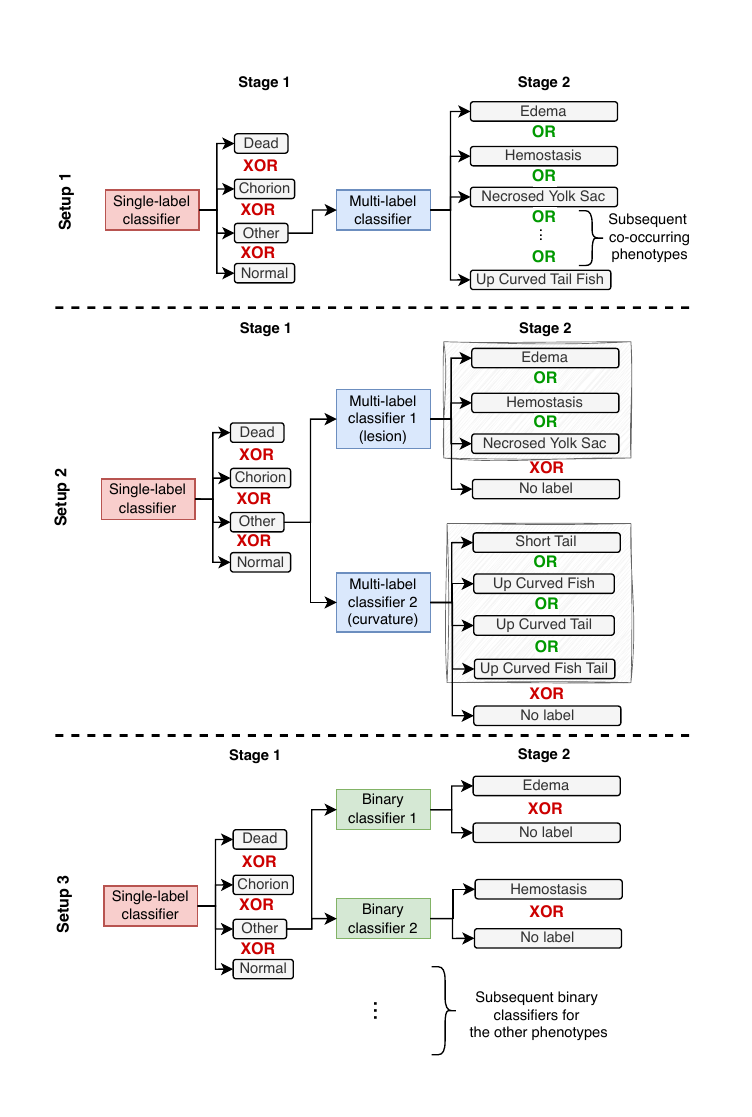}}
\caption{Three experimental ensemble setups are evaluated in this study. XOR denotes an exclusive alternative, i.e., \textbf{exactly one} of the possible options is selected, whereas OR denotes a non-exclusive alternative, in which \textbf{at least one} of the available options is selected.}
\label{fig:setups}
\end{figure}

The three image recognition models introduced in Section~\ref{sec:Intro}, a CNN based on ResNet18, a Vision Transformer (ViT), and a ConvNeXt model, were evaluated independently within the three experimental setups illustrated in Fig.~\ref{fig:setups}. All setups share an identical first stage, in which a single-label classifier assigns each image to one of four categories: Dead, Chorion, Normal, or Other. Only images classified as Other proceed to the second stage for further analysis. The second stage differs across the three setups and is defined as follows:
\begin{itemize}
    \item \textbf{Setup 1} --- A single multi-label classifier is used to assign phenotypes in the second stage. This classifier operates in a non-exclusive (OR) mode, allowing multiple phenotypes to be assigned to a single image. A phenotype is predicted if its associated probability exceeds a predefined threshold. The procedure for tuning this threshold using the validation set is described below.

    \item \textbf{Setup 2} --- The second stage consists of two specialized multi-label classifiers. The first classifier focuses on \textit{lesion-related} phenotypes and can assign at least one of the following labels: Edema, Hemostasis, or Necrosed Yolk Sac, or no label. It is explicitly trained to produce no label using images labeled as Normal. The second classifier targets morphological changes related to \textit{tail or body curvature} and can assign at least one of the following labels: Short Tail, Up Curved Fish, Up Curved Tail, or Up Curved TailFish, or no label. As previously, it is trained to output no labels using images with the Normal phenotype.

    \item \textbf{Setup 3} --- In this setup, the second stage comprises a set of independent binary classifiers, one for each phenotype. Each classifier determines the presence or absence of its corresponding phenotype. As in setup 2, images labeled as Normal are used to train the classifiers to produce no positive predictions.
\end{itemize}

In each setup, the same model architecture (ResNet18, ViT, or ConvNeXt) is employed consistently across both stages. All models are loaded using the \textit{timm} Python library with pretrained weights. Specifically, the architectures version employed by us are: \textit{resnet18}, \textit{vit\_tiny\_patch16\_224}, \textit{convnext\_tiny}.

During training, the validation dataset serves two purposes. First, it is used to monitor the loss function throughout the training process. Second, in the second-stage classification, it is used to determine the optimal probability thresholds for assigning individual phenotypes. Specifically, each multi-label or binary classifier outputs the probability of the presence of a given phenotype. For each phenotype, the decision threshold is selected from a predefined set of values, $[0.05, 0.10, 0.15, \ldots, 0.95]$, so as to maximize the F1-score on the corresponding validation subset.

We maintain the same training parameters for all setups and architectures. Specifically, the batch size is set to 16, learning rate is 0.0001, numbers of epochs in both stages are 10, and weight decay equals 0.0001. 

The evaluation of the performance of each setup and each architecture is conducted using three main measures: precision, recall, and F1-score. We differ in this choice from Jeanray et al. \cite{Jeanray2015-phenotype} which in their study used mainly recall (identification) as the classification quality measure. For a phenotype class \textit{c}, we define:
\begin{itemize}
    \item \textbf{True Positive, TP(c)} - number of images whose true label is \textit{c} and whose predicted label set contains \textit{c}.
    \item \textbf{False Negative, FN(c)} -  number of images whose true label is \textit{c} and whose predicted label set does not contain \textit{c}.
    \item \textbf{False Positive, FP(c)} - number of images whose true label is not \textit{c}, but whose predicted label set contains \textit{c}.
\end{itemize}

Thus,
\begin{gather*}
precision(c) = \frac{TP(c)}{TP(c) + FP(c)}, \qquad
recall(c) = \frac{TP(c)}{TP(c) + FN(c)},\\
F1(c) = \frac{2 \cdot precision(c) \cdot recall(c)}{precision(c) + recall(c)},\\
accuracy = \frac{TP + TN}{TP + TN + FP + FN}.
\end{gather*}

\section{Results and Discussion of Experiments}
\label{sec:Results}

In Table~\ref{Table:meancompare}, we present a comparison of the mean precision, recall, and F1-score across all evaluated setups and architectures. These metrics are averaged over all phenotypes on the test dataset, and the best value for each measure is highlighted in bold. Several key observations can be drawn from the results in Table~\ref{Table:meancompare}.

\begin{itemize}
    \item Overall, ConvNeXt consistently achieves the best performance across all three setups. This is a notable finding, as the ConvNeXt architecture has not been previously explored in the context of zebrafish phenotype classification.
    
    \item Interestingly, setup~2 yields the highest mean precision and F1-score, while also achieving the second-highest recall (for ConvNeXt). To the best of our knowledge, such a specialised ensemble configuration has not been previously investigated in the literature. In contrast, prior studies (e.g., Jeanray et al.~\cite{Jeanray2015-phenotype} and Tandon et al.~\cite{Tandon2025-AIToxicityZebrafish}) mainly employ approaches corresponding to setup~3. Notably, the highest recall overall is obtained by ConvNeXt in setup~3. Similarly, ViT also achieves best precision and recall for setup~2.
    
    \item All setups exhibit a tendency toward higher recall than precision. The largest disparity between these metrics is observed for ResNet18 in setup~1, indicating a strong bias toward recall in this configuration.
\end{itemize}

\begin{table}[H]
	\centering
	\caption{Comparison of the mean precision, recall, F1-score, and accuracy aggregated over all phenotypes for the evaluated setups and model architectures (m. stands for mean).}
    \label{Table:meancompare}
    \resizebox{\textwidth}{!}{%
    \begin{tabular}{cccccccccc}
		\toprule
		&\multicolumn{3}{c}{\textbf{Setup 1}}&\multicolumn{3}{c}{\textbf{Setup 2}}&\multicolumn{3}{c}{\textbf{Setup 3}}\\
		\cmidrule(lr){2-4}\cmidrule(lr){5-7}\cmidrule(lr){8-10}
		\textbf{Measure}  &  ResNet18 & ViT &  ConvNeXt & ResNet18 & ViT & ConvNeXt &  ResNet18 & ViT & ConvNeXt \\
		\cmidrule{1-10}
m. precision & 0.32 & 0.33 & \textbf{0.40} & 0.35 & 0.38 & \textbf{0.40} & 0.33 & 0.34 & 0.36 \\
m. recall    & 0.85 & 0.46 & 0.53 & 0.61 & 0.68 & 0.73 & 0.69 & 0.71 & \textbf{0.87} \\
m. F1-score  & 0.38 & 0.32 & 0.42 & 0.40 & 0.45 & \textbf{0.47} & 0.41 & 0.41 & 0.44 \\
accuracy & \textbf{0.85} & 0.53 & 0.55 & 0.61 & 0.74 & 0.76 & 0.72 & 0.79 & 0.84 \\
	\bottomrule
\end{tabular}}
\end{table}

In terms of achieved accuracy, the highest accuracy was achieved by setup 1 with ResNet18 at 0.85, followed closely by setup 3 with ConvNeXt at 0.84. Across setups, setup~3 was the most consistent, with all models above 0.71, while setup~1 showed the largest spread between ResNet18 and the other two models. Thus, setup 3 achieved the most consistent and highest accuracy, mainly due to stronger recall, but setup 2 remained superior in terms of F1-score, indicating a better balance between precision and recall.

\begin{figure}
\centerline{\includegraphics[width=0.9\textwidth]{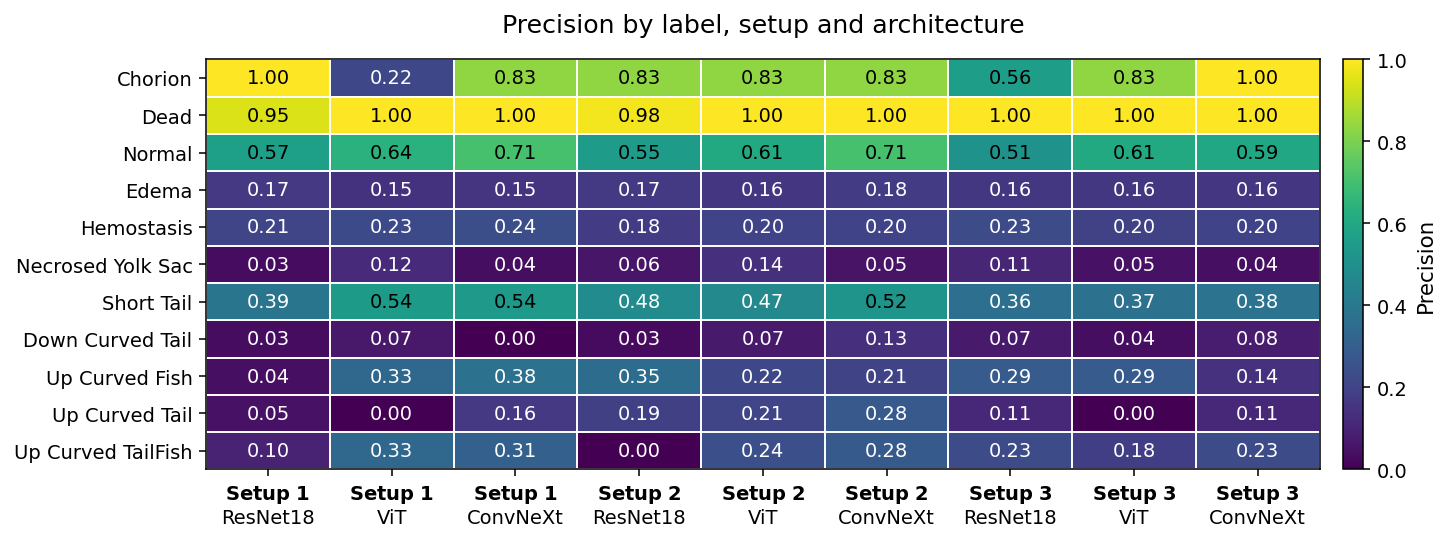}}
\centerline{\includegraphics[width=0.9\textwidth]{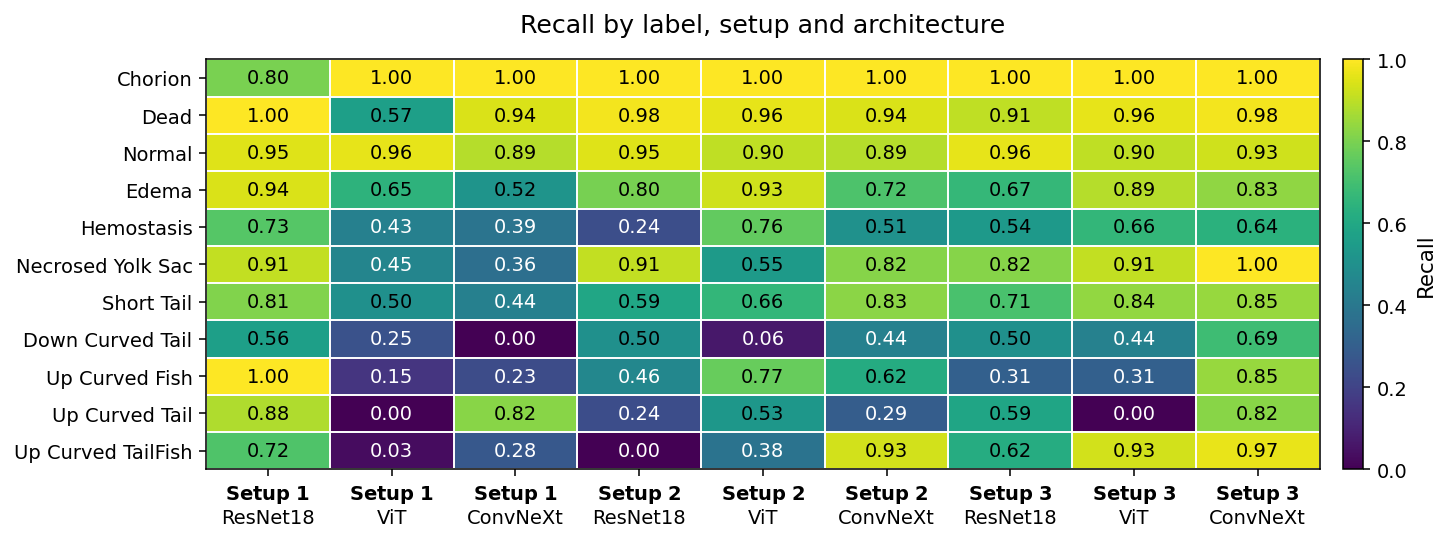}}
\centerline{\includegraphics[width=0.9\textwidth]{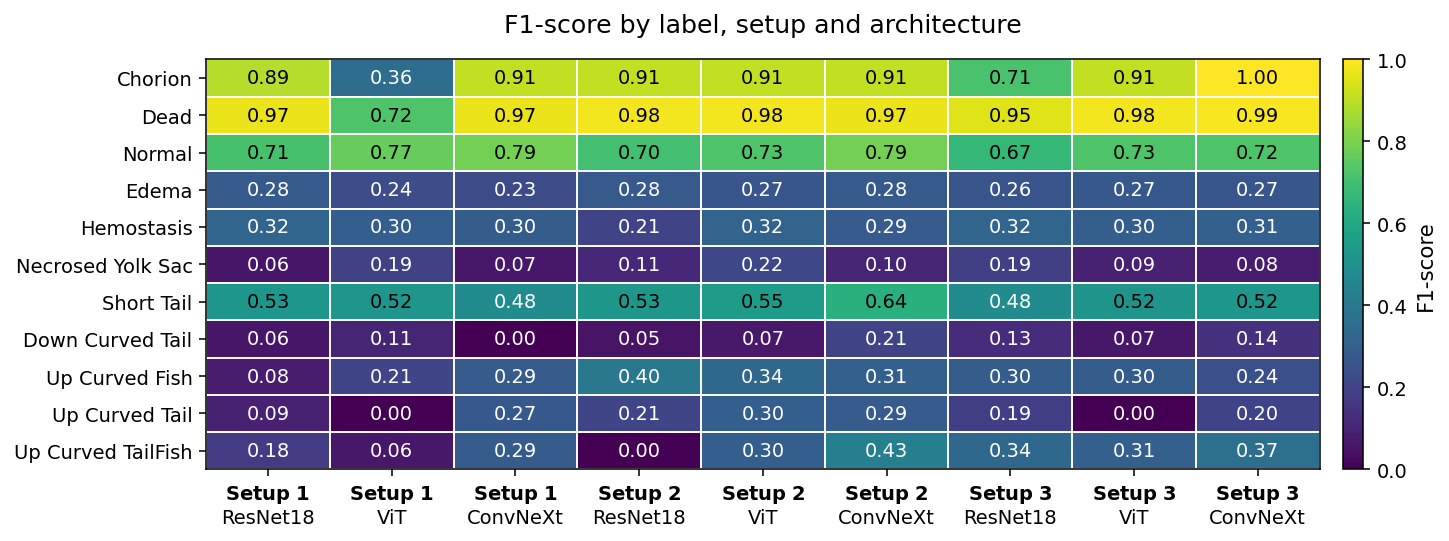}}
\caption{Heatmaps of precision, recall and F1-score calculated over testing dataset for each phenotype, setup and architecture.}
\label{fig:measures-heatmaps}
\end{figure}

We further analyse precision, recall, and F1-score at the level of individual phenotypes, setups, and architectures. The corresponding results are visualised as heatmaps in Fig.~\ref{fig:measures-heatmaps}.

\begin{itemize}
    \item For the three mutually exclusive classes---Dead, Chorion, and Normal---the values of precision, recall, and F1-score are consistently high across all evaluated models.
    
    \item For the remaining phenotypes (i.e., those that may co-occur within the same images), recall is generally higher than precision. Among these, the best classification performance is observed for the Short Tail phenotype, particularly in setup~2 when using the ConvNeXt architecture.
    
    \item The results also suggest that certain models, such as ViT, are more effective at identifying more challenging phenotypes. For example, for the Hemostasis phenotype, ViT consistently achieves higher recall across all three setups compared to ResNet18 and ConvNeXt.

    \item ConvNeXt frequently achieves the highest recall values for complex phenotypes (e.g., Short Tail, Up Curved TailFish, Down Curved Tail in setup~3), suggesting that it is particularly effective at detecting subtle morphological variations. However, this often comes at the cost of reduced precision.

    \item As indicated previously, ConvNeXt in setup~2 in overall yields the best F1-score across most classes. In our opinion, building specialised ensembles are presented in setup~2 tends to provide better overall results than setup~3 which was presented in \cite{Jeanray2015-phenotype}.
\end{itemize}

\begin{figure}[h!t]
    \centering
    \includegraphics[width=1.2\linewidth]{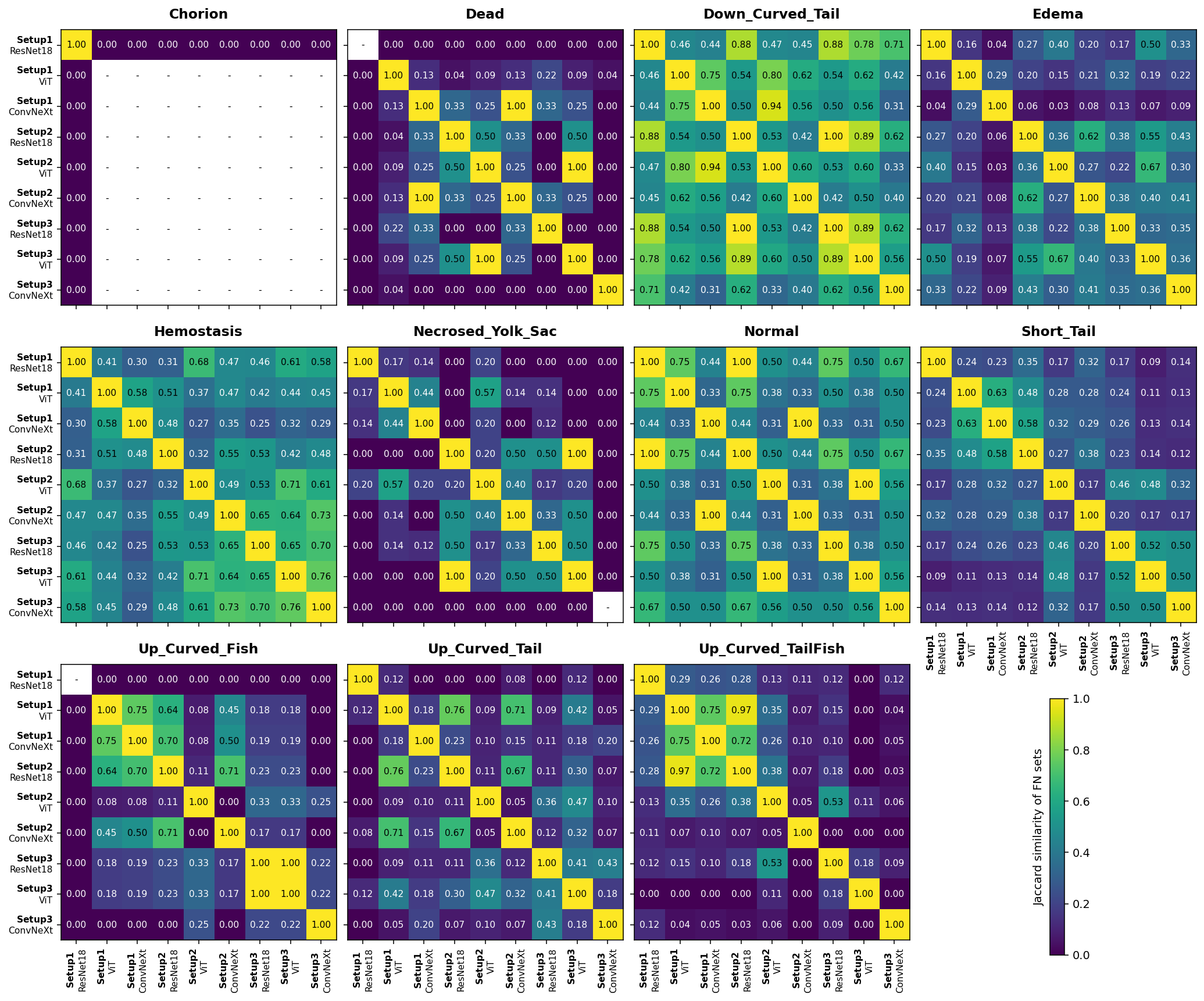}
    \caption{Heatmaps of Jaccard similarity for cross-setup and cross-model comparisons of False Negative sets $FN(c)$ for each phenotype $c$. The symbol (-) indicates that at least one model produced no False Negatives for a given class (i.e., the similarity is undefined), whereas a 0 value indicates no overlap between the corresponding $FN$ sets.}
    \label{fig:Jaccard}
\end{figure}

In Fig.~\ref{fig:Jaccard}, we further analyze the results with a focus on \textit{False Negatives}, $FN(c)$. In particular, we present cross-setup and cross-model heatmaps illustrating the \textit{Jaccard similarity} between the sets of False Negatives for each class.

For a given class $c$ and two models or setups $A$ and $B$, the Jaccard similarity is defined as:
\[
Jacc\big(FN(c)_{A}, FN(c)_{B}\big) = \frac{|FN(c)_{A} \cap FN(c)_{B}|}{|FN(c)_{A} \cup FN(c)_{B}|}.
\]

The Jaccard similarity quantifies the extent to which False Negatives are shared between models and setups. For example,
\begin{multline*}
Jacc\big(FN(\text{Hemostasis})_{\text{Setup1-ResNet18}},\\
FN(\text{Hemostasis})_{\text{Setup1-ConvNeXt}}\big) = 0.3
\end{multline*}
indicates that $30\%$ of the combined False Negative samples for the \textit{Hemostasis} class are shared between the ResNet18 and ConvNeXt models within setup~1.

The results reveal that the degree of overlap varies across classes. For some phenotypes, the overlap is relatively low, suggesting model-specific errors, whereas for others it is higher, indicating that different models tend to miss the same samples. Notably, higher Jaccard values are more frequently observed in setup~3, suggesting greater consistency of False Negatives across models in this configuration compared to the other setups.

\section{Conclusions}
\label{sec:Conclusions}

In this article, we tested three hierarchical ensemble setups to classification of zebrafish phenotypes. Each setup consists of two stages: the first stage is the same for each setup and its aim is to use a single classifier to classify zebrafish embryo images into one of four exclusive phenotypes: Normal, Chorion, Dead or Other. Only images classifies into the Other state proceed to the second stage. In stage 2, the ensemble models in each setup are organized differently, however they all perform multi-label classification of images. 
\begin{itemize}
    \item In the first setup, there is a single multi-label classifier which aim to assign to a classified image all relevant labels of detected deficiencies: Edema, Hemostasis, or curvature defects. 
 
    \item In the second setup, we tested two separate models in stage 2. The first classifier focuses on \textit{lesion-related} phenotypes and the second classifier targets morphological changes related to \textit{tail or body curvature}.

    \item In the third setup there is an ensemble of binary classifiers, each making a separate detection of a corresponding phenotype in each image. 
\end{itemize}
In all three setups we tested three backbone image recognition architectures: ResNet18, ViT and ConvNeXt, from which ConvNeXt was not previously applied to image recognition in zebrafish embryo images. 

In all tested setups and backbone architectures, generally ConvNeXt consistently achieves the best performance. Interestingly, setup~2 yields the highest mean precision and F1-score, while also achieving the second-highest recall (for ConvNeXt). To the best of our knowledge, such a specialised ensemble configuration has not been previously investigated in the literature. Generally, all ensembles perform better in terms of recall than precision. The high achieved recall values are inline with those presented by Jeanray et al.~\cite{Jeanray2015-phenotype}. 

Thus, the two main conclusions of our work are:
\begin{enumerate}
    \item Overall, the ConvNeXt architecture yields the best classification results, even for the most difficult phenotypes to classify, such as Downed Curved Tail.
    
    \item The proposed hierarchical specialised ensemble (setup~2) tends to perform best in terms of F1-score among three tested setups.
\end{enumerate}

Particular phenotypes such as downward- and upward-curved tails were associated with both lower precision and recall, indicating that these features are particularly sensitive and challenging to classify accurately. This suggests that incorporating quantitative descriptors in the future work, such as tail length and curvature metrics, could improve the assessment of phenotype severity and enhance model performance in future studies. Furthermore, the inclusion of additional morphological features describing eye size, yolk sac edema, and internal organ malformations, could provide a more comprehensive evaluation of toxicological and teratogenic effects. 

\section*{The use of AI tools}
The LLM model (Codex) was used to polish writing of the manuscript, mainly including correction of grammar mistakes, and for supporting the preparation of the setups and models implementations used in the experiments.
\section*{Acknowledgements}
The first author acknowledges support by the Warsaw University of Technology Research University - Excellence Initiative program [Grant numbers CPR-IDUB/288/Z01/POB3/2024 and CPR-IDUB/504/04496/1032/45.000027].

\bibliographystyle{splncs04}
\bibliography{my-references}

\end{document}